\documentclass[10pt,twocolumn,letterpaper]{article}
\usepackage{xspace}
\newcommand{\method}{DDP\xspace}
\usepackage{cvpr}              %
\usepackage{booktabs} %
\usepackage{multirow}
\usepackage{marvosym}

\usepackage{indentfirst}

\definecolor{cvprblue}{rgb}{0.21,0.49,0.74}
\usepackage[pagebackref,breaklinks,colorlinks,allcolors=cvprblue]{hyperref}

\def\paperID{5} %
\def\confName{CVPR}
\def\confYear{2026}

\title{Less Detail, Better Answers: Degradation-Driven Prompting for VQA}

\author{
    Haoxuan Han\textsuperscript{*}\quad
    Weijie Wang\textsuperscript{*}\textsuperscript{\dag}\quad
    Zeyu Zhang\quad
    Yefei He\quad
    Bohan Zhuang\textsuperscript{\Letter}  \\[1em]
    State Key Lab of CAD\&CG, Zhejiang University
}

\begin{document}
\pagestyle{empty}  %
\maketitle

\let\thefootnote\relax\footnotetext{\textsuperscript{*} Equal contribution. \textsuperscript{\dag} Project Lead: Weijie Wang (wangweijie@zju.edu.cn). \textsuperscript{\Letter} Corresponding author: Bohan Zhuang (bohan.zhuang@zju.edu.cn).}

\begin{abstract}
Recent advancements in Vision-Language Models (VLMs) have significantly pushed the boundaries of Visual Question Answering VQA. However, high-resolution details can sometimes become \textit{noise} that leads to hallucinations or reasoning errors. In this paper, we propose Degradation-Driven Prompting (DDP), a novel framework that improves VQA performance by strategically reducing image fidelity to force models to focus on essential structural information. We evaluate DDP across two distinct tasks. Physical attributes targets images prone to human misjudgment, where DDP employs a combination of 80p downsampling, structural visual aids white background masks and  orthometric lines, and In-Context Learning ICL to calibrate the model’s focus. Perceptual phenomena addresses various machine-susceptible visual anomalies and illusions, including Visual Anomaly VA, Color CI, Motion MI, Gestalt GI, Geometric GSI, and Visual Illusions VI. For this task, DDP integrates a task-classification stage with specialized tools such as blur masks and contrast enhancement alongside downsampling. Our experimental results demonstrate that \textit{less is more} by intentionally degrading visual inputs and providing targeted structural prompts, DDP enables VLMs to bypass distracting textures and achieve superior reasoning accuracy on challenging visual benchmarks. Code is  available on our project page: \url{https://hhx-jpg.github.io/ddp/}.
\end{abstract}

\section{Introduction}
\label{sec:intro}

\begin{figure}[t]
\begin{center}
    \includegraphics[width=\linewidth]{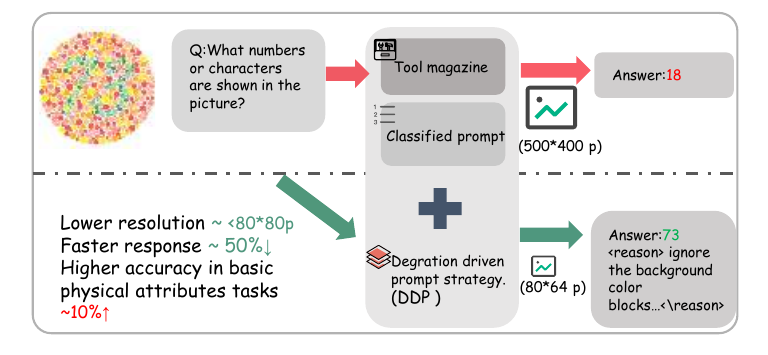}
\end{center}
\vspace{-0.05cm}
\caption{\textbf{Less is More in Visual Perception.} Comparison between a standard high-resolution pipeline and our proposed DDP strategy. While high-resolution inputs ($500 \times 400$p) can paradoxically lead to misinterpretation (e.g., identifying "18" instead of "73"), our DDP leverages lower resolution ($80 \times 64$p) to eliminate background noise. This approach achieves a ~50\% reduction in response time and a ~50\% improvement in accuracy for basic physical attribute tasks by focusing the model on essential structural information.}
\label{fig:teasor}
\end{figure}

Vision-Language Models (VLMs) \cite{geminiteam2024gemini,gao2023gllava,openai2024gpt4o} have demonstrated remarkable capabilities in cross-modal understanding, image captioning, and complex visual reasoning.
Despite achieving state-of-the-art results on standard benchmarks, these models often exhibit a surprising lack of robustness when faced with visually deceptive images, such as optical illusions.
This vulnerability suggests that current scaling laws, while effective for general recognition, may be hitting a plateau regarding deep structural understanding.
As highlighted by \cite{gao2025pixelspatternspoetryworld}, there is a fundamental discrepancy between how machines and humans perceive the world; while humans possess a poetic ability to generalize global structures and semantic logic, VLMs remain heavily reliant on local textures and statistical patterns pixels and patterns.
This limitation makes them highly susceptible to visual traps in which local features contradict global reality, leading to confident but catastrophically incorrect hallucinations.

The core challenge lies in the passive nature of the current VLM inference.
Standard models typically employ a single-shot observation approach akin to a fast, instinctive  response which is often insufficient to resolve the ambiguity inherent in deceptive visual signals.
To bridge this gap, we argue that a shift from passive observation to agentic perception\cite{deng2023mind2web,jia2022visualprompttuning,park2026vagentinteractivevideosearch} is required.
This shift mimics the human cognitive process of slow thinking, where initial visual cues are scrutinized through deliberate focus.
Drawing inspiration from DeepEyesV2 ~\cite{hong2025deepeyesv2agenticmultimodalmodel}, we posit that VLMs should not merely be static evaluators but active agents capable of interacting with their environment.
By utilizing tool-use and iterative verification, these models can refine their initial, often flawed, visual intuitions into a grounded understanding of the scene's underlying geometry.
\begin{figure*}[t]
\begin{center}
    \includegraphics[width=\textwidth]{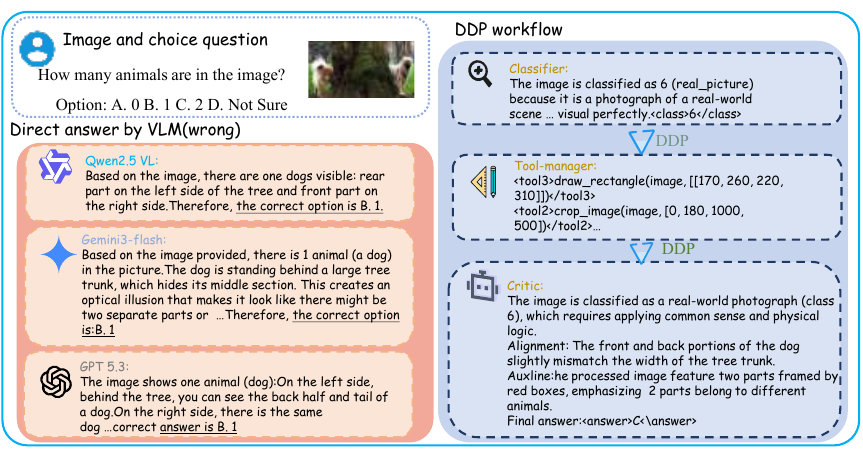}
\end{center}
\vspace{-0.05cm}
\caption{\textbf{Overcoming visual reasoning bottlenecks via the DDP framework.} 
Standard VLMs are easily deceived by optical illusions or occlusions (e.g., a dog seemingly split by a tree). 
Our DDP approach introduces a "divide-and-conquer" strategy: the \textbf{classifier} categorizes the image type, the \textbf{tool manager} invokes specialized visual tools (e.g., \textit{draw\_rectangle} and \textit{crop}) to highlight suspicious regions, and the \textbf{Critic} synthesizes these visual cues. 
The Critic detects that the front and back portions mismatch the trunk width, thereby correcting the initial misconception and providing the right answer (\textit{Option C: 2}).}
\label{fig:teasor}
\end{figure*}

In this paper, we propose \method, a multi-faceted framework designed to enhance VLM performance on challenging visual tasks by integrating downsampling strategies, advanced prompt engineering, and agentic tool-use.
Our approach is motivated by three key insights

Structural Over-Sensitivity. To mitigate the model's over-reliance on deceptive high-frequency textures, we introduce the DDP mechanism.
By reducing local noise and fine details, we effectively create a low-pass filter that forces the model to prioritize global topological structures.
This simulates the human tendency to squint or step back to resolve optical ambiguities that disappear when high-frequency information is suppressed.
Fine-Grained Interaction. Building upon the concepts in Fine-Grained Visual Prompting ~\cite{yang2023finegrainedvisualprompting}, we utilize targeted visual prompts to guide the model's attention toward specific contradictory regions within an illusion.

Agentic Tool Augmentation. Following the DeepEyesV2 training philosophy ~\cite{hong2025deepeyesv2agenticmultimodalmodel}, we empower the VLM to call on external diagnostic tools such as blurmasks, grid auxlines, and localized magnifiers.
By treating the VLM as an agent that can test its visual hypotheses against objective geometric data, we move closer to the robust generalization and self-correction capabilities observed in human vision.
The primary contributions of this work are as follows:
 We provide a systematic analysis of VLM vulnerabilities in the context of visual illusions, framing the problem through the lens of the perception-logic gap\cite{gao2025pixelspatternspoetryworld} and establishing a new set of benchmarks for deceptive vision.
We introduce a novel inference pipeline that combines multi-resolution downsampling and fine-grained visual prompting to improve the structural awareness of multimodal models.
An agentic framework is implemented where the VLM actively utilizes external tools to verify visual inputs, significantly reducing hallucination rates.
 Experimental results demonstrate that our collaborative approach consistently outperforms traditional end-to-end VLM inference, providing a scalable and interpretable pathway toward more reliable and human-like machine vision.

\section{Related Works}
\begin{figure*}[t]
    \centering
    \includegraphics[width=\textwidth]{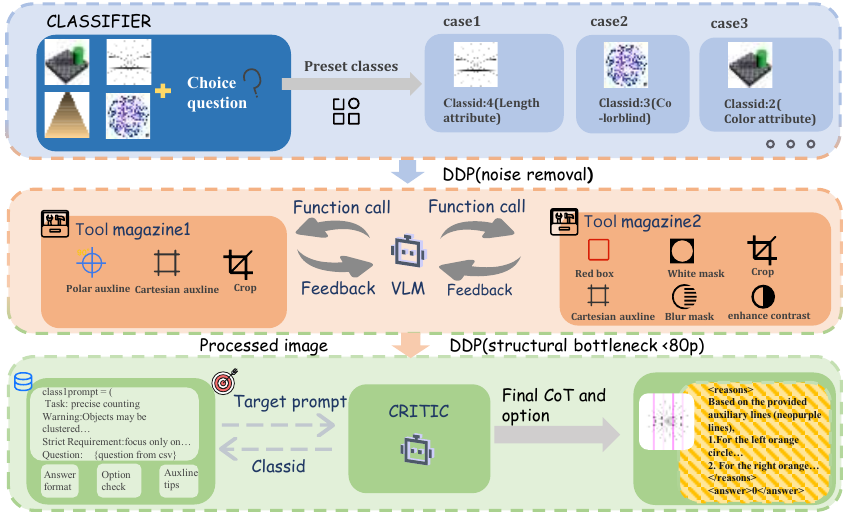}

        \caption{\noindent\textbf{Overview of the DDP-based VLM enhancement framework.} 
The workflow consists of three primary stages: 
(1) \noindent\noindent\textbf{classifier}: The input image and choice question are categorized into preset domains (e.g., motion illustration, colorblindness, or color attributes) to guide subsequent tool selection. 
(2) \noindent\textbf{tool manager}: After an initial noise-removal downsampling, a DDP agent performs iterative function calls to select specialized visual tools from magazines, such as \textit{Polar/Cartesian auxiliary lines}, \textit{crops}, and \textit{masks}, to highlight key features. 
(3) \noindent\textbf{target prompting}: Despite the extreme downsampling (structural bottleneck $<$ 80p), the processed image is fed into the Critic module. By leveraging task-specific prompts and auxiliary tips based on the ClassID, the system generates a structured Chain-of-Thought (CoT) reasoning process to produce the final option.}
    
    \label{fig:pipeline}
    
\end{figure*}

\subsection{Agentic Multimodal Model}
The paradigm shift from passive observation to \textit{agentic perception} is characterized by models that do not merely generate descriptions but actively manipulate their cognitive processes and environment.
Visual Programming and Tool Augmented Reasoning. Early efforts like VisProg \cite{gupta2023visual} and ViperGPT \cite{suris2023vipergptvisualinferencepython} pioneered code generation for visual tasks enabling complex reasoning through modular tool use while Chameleon \cite{lu2023chameleon} integrates external tools to augment LLM knowledge. Similarly HuggingGPT \cite{shen2024hugginggpt} and Visual ChatGPT \cite{wu2023visual} orchestrate task specific models showcasing multi tool potential.
Active Perception and Fine grained Interaction. Addressing the \textit{perception logic gap} recent research emphasizes active analysis where Ferret \cite{you2023ferret} enables fine grained grounding and models like LLaVA Interactive \cite{chen2023llavai} and LLaVA Plus \cite{liu2023llava} incorporate visual prompts for real world editing. Furthermore SPHINX \cite{lin2023sphinxjointmixingweights} and InternLM XComposer2 \cite{dong2024internlm} support multi resolution inputs allowing models to zoom in on critical regions.
Multimodal Agents in Complex Environments. In dynamic environments AppAgent \cite{yang2023appagent} and Mobile Agent \cite{wang2024mobile} operate smartphones via visual feedback while Mind2Web V \cite{deng2023mind2web} and SeeClick \cite{cheng2024seeclick} focus on web navigation. Cradle \cite{tan2024cradleempoweringfoundationagents} explores gaming agents and VILA Agent \cite{xu2024vilaagent} with Agent V \cite{park2026vagentinteractivevideosearch} target long horizon planning and self correction.
Benchmarks for Agentic Evaluation. For evaluation benchmarks like MathVerse \cite{zhang2024mathverse} and MMSearch \cite{jiang2024mmsearchbenchmarkingpotentiallarge} assess logic whereas ToolBench \cite{qin2023toolllmfacilitatinglargelanguage} and RealX Bench measure tool invocation capabilities bridging the gap between reasoning and perception.

\subsection{Visual Prompting for VLMs}
The evolution of visual prompting has transitioned from simple indicators to semantic overlays.
Marker based grounding and set of mark. A key milestone is Set of Mark SoM \cite{yang2023setofmarkpromptingunleashesextraordinary} which uses alphanumeric symbols for region grounding a method extended by Ferret \cite{you2023ferret} and GLaMM \cite{rasheed2024glamm} to integrate continuous spatial coordinates for interacting with arbitrary shapes.
Interactive and Point based Prompting. To facilitate natural interaction ViP LLaVA \cite{cai2024vipllavamakinglargemultimodal} and Draw and Understand \cite{chen2024draw} process arbitrary cues like scribbles while Shtedritski et al \cite{shtedritski2023what} demonstrate that simple markers act as \textit{visual anchors} guiding cross attention networks.
Multi scale and grid based Strategies. Addressing high resolution scenes LLaVA UHD \cite{xu2024llavauhd} and Monkey \cite{li2023monkey} utilize grid based cropping strategies whereas V IRL \cite{hu2024virl} and G LLaVA \cite{gao2023gllava} incorporate geometric annotations for spatial navigation and problem solving.
In context Visual Learning. Finally \textit{In context Visual Learning} methods such as Visual Prompt Tuning VPT \cite{jia2022visualprompttuning} and Images Speak in Images \cite{wang2023imagesspeakimagesgeneralist} use visual demonstrations to steer behavior while Mantis \cite{jiang2024mantis} leverages interleaved contexts for comparative reasoning.

The evolution of visual prompting has moved from basic indicators toward sophisticated semantic overlays that bridge the gap between raw pixel data and linguistic instructions. A major development in this area involves the use of alphanumeric symbols and markers to establish region grounding which allows models to associate specific parts of an image with text descriptions.These visual anchors serve as critical guidance for cross attention mechanisms helping the model focus its computational resources on the most relevant areas for answering questions or performing tasks.

As research progresses toward handling high resolution scenes and complex reasoning strategies have become essential for modern architectures.  Parallel to these structural improvements in context visual learning has emerged as a powerful paradigm for steering model behavior through visual demonstrations. Methods like visual prompt tuning and the images speak in images approach provide the model with examples that clarify the desired output format or reasoning logic. More recent efforts such as mantis leverage interleaved contexts to support comparative reasoning across multiple images which allows for a deeper understanding of relationships within a visual sequence or a collection of related scenes.

\section{Methodology}

In this section, we detail our proposed multi-stage inference pipeline designed to mitigate visual illusions in Vision-Language Models. The framework consists of three primary stages: initial perception and hierarchical task classification, agentic tool invocation for feature disentanglement, and low-resolution refinement with targeted prompting.

\subsection{Overview}
The core observation motivating our Degradation-Driven Prompting (DDP) is that excessive visual detail in high-resolution images often acts as a distraction, diverting VLMs from the main task and significantly increasing context length. DDP is not the entire pipeline but refers specifically to the strategic injection of downsampling operations at key stages to enforce structural attention. By employing these degradation steps, we not only improve computational efficiency but also force the VLM's attention mechanism to focus on the essential structural elements required for the task.

We detail our proposed multi-stage inference pipeline designed to mitigate visual illusions in Vision-Language Models. The framework consists of three primary stages,please refer to ~\cref{sec3.2,sec3.3,sec3.4}. The DDP mechanism is applied at the beginning of ~\cref{sec3.3} and ~\cref{sec3.4} respectively.

\subsection{Task Classification}\label{sec3.2}
The \method begins with a task classification stage. To mitigate the impact of high-frequency deceptive signals at the outset, the input image $I$ is first subjected to a light-weight Gaussian smoothing $\sigma_1$ to generate a base perception:
\begin{equation}
I_{base} = \text{Smooth}(I, \sigma_1)
\end{equation}
The classifier then performs a dual-level taxonomy analysis to determine the routing logic for subsequent stages. Given the smoothed image and the user query $Q$, the classifier produces a task configuration $\mathcal{C}$:
\begin{equation}
\mathcal{C} = \text{classifer}(I_{base}, Q)
\end{equation}
We categorize the visual challenges into two primary task sets based on their cognitive complexity, which are further divided into specific subcategories:
\begin{itemize}
    \item \noindent\noindent\textbf{Physical Attributes:} Focuses on objective geometric and photometric properties. This major category is subdivided into \textit{Size}, \textit{Length}, \textit{Color}, and \textit{Others}.
    \item \noindent\textbf{Perceptual Phenomena:} Addresses higher-level cognitive illusions that deceive the  global reasoning of VLM. This category is subdivided into eight specific sub-tasks: \textit{Counting}, \textit{Find Difference}, \textit{Color-Blind}, \textit{Motion}, \textit{Geometry}, \textit{Real Picture}, \textit{Size}, and \textit{Others}.
\end{itemize}
The configuration $\mathcal{C}$ dictates the selection of specialized visual primitives in the next stage.

\subsection{Tool Manager}\label{sec3.3}
In this stage, we apply the first level of Degradation-Driven Prompting (DDP), where the input image is preliminarily compressed to a maximum dimension (height or width) of approximately 150 pixels ($I_{DDP}$) to filter out high-frequency noise while retaining sufficient detail for tool application.
\begin{equation}
I_{DDP} = \text{Downsample}(I_{base}, R_{150})
\end{equation}
Upon receiving the task category $\mathcal{C}$ from the classifer, the Tool Manager functions as an autonomous agent designed to decouple deceptive visual signals by invoking specialized external tools $\mathcal{T}$ from a predefined library $\Omega$. The core objective of this stage is to transform the ambiguous or illusion-contaminated input into a series of evidence-enhanced representations that facilitate objective reasoning. The interaction is formulated as the generation of a tool-processed image set:
\begin{equation}
I_{tool} = \mathcal{T}(I_{DDP}, \theta)
\end{equation}
In this equation, $\theta$ represents the execution parameters, such as position of aux lines,  angles of aux lines, which are dynamically predicted by the agent based on the semantic requirements of the task.

\noindent\textbf{Tools for Physical Attribute Decoupling.}
For tasks categorized under Physical Attributes, such as assessing geometric proportions or rectifying perspective distortions, the Tool Manager employs a suite of auxiliary lines and spatial-isolation primitives. The polar aux line is utilized to serve as the reference to eliminate the illusion of distortion of slant lines, effectively serving as a reference to whether two lines are aligned.
 The cartesian aux line (or perspective rectification) is applied to correct skewed planes, ensuring that parallel lines in 3D space are projected accurately for the evaluation of critic~\cref{sec3.4}. To eliminate peripheral distractions and context-induced size illusions (e.g., the Ebbinghaus illusion), the crop tool is invoked to extract the target objects at a uniform scale:
\begin{equation}
I_{crop} = \text{Extract}(I_{DDP}, [x, y, w, h])
\end{equation}
These transformations ensure that the physical dimensions of the stimuli are presented in a canonical, measurement-ready format.

\noindent\textbf{Tools for Perceptual Phenomenon Mitigation.}
To mitigate higher-level cognitive illusions that rely on the "Gestalt" effect or deceptive local textures, the tool manager deploys a complex array of "signal-disturbing" tools. The red box and white-out masking tools are used to break global context; the former forces the model’s spatial attention onto a specific boundary, while the latter replaces the surrounding deceptive gradients with a neutral constant $\mathbf{1}_{white}$ to isolate the target's true luminance or color:
\begin{equation}
I_{mask} = I_{DDP} \odot M + (1 - M) \cdot \mathbf{1}_{white}
\end{equation}
For alignment and parallelism verification in structural illusions (e.g., the Poggendorff illusion), the cartesian aux line tool overlays specific orthogonal reference lines,horizontal or vertical at critical intersection points predicted by the agent. 
\begin{equation}
I_{aux} = \text{Overlay}(I_{DDP}, L_{\{hori, vert\}})
\end{equation}
Additionally, blur masking is applied to suppress high-frequency deceptive textures that often trigger motion or depth illusions in VLMs. By convolving the image with a heavy Gaussian kernel $G(\sigma)$, the tool forces the model to ignore noisy textural cues and focus on stable, low-frequency structural signals:
\begin{equation}
I_{blur} = I_{DDP} \times G(\sigma), \quad \sigma \gg \sigma_1
\end{equation}
Finally, the crop tool is re-utilized here to provide a "foveated" view of disputed regions, ensuring that the critic's synthesis is based on purified, localized evidence rather than the original deceptive global configuration.

\subsection{Target Prompting}\label{sec3.4}
Standard VLMs are known to be heavily reliant on local textures, often prioritizing high-frequency details over global semantics. To mitigate this high-frequency bias , we implement the most aggressive form of DDP in this final stage, where the image is further downsampled to a maximum dimension of 80 pixels . This extreme reduction acts as a structural bottleneck, effectively stripping away deceptive local textures and forcing the model to rely solely on global structural cues.
During this stage, the critic is responsible for synthesizing the raw image, the tool-augmented evidence, and the task-specific logic to produce a calibrated judgment. Unlike standard VLMs that may fall into the trap of over-interpreting pixel-level noise, our critic avoids this pitfall by operating through the aforementioned structural bottleneck.

\noindent\textbf{Structural Bottleneck via Strategic Downsampling}
We introduce a radical downsampling $\mathcal{B}$ that reduces the tool-processed image $I_{tool}$ to an extreme low-resolution representation $I_{DDP}$ (typically $R \leq 80p$):
\begin{equation}
I_{DDP} = \text{Downsample}(I_{tool}, R_{80})
\end{equation}
The theoretical motivation is grounded in the \textit{Data Processing Inequality}. By deliberately discarding high-resolution details, we minimize the mutual information between the deceptive textural noise $N$ and the final prediction $Y$:
\begin{equation}
\mathcal{I}(N; Y) \to 0
\end{equation}
This bottleneck forces the critic to focus exclusively on the topological and structural alignment of the objects.

\noindent\textbf{Evidence Synthesis and Final Alignment.}
The critic receives the low-resolution "purified" image $I_{DDP}$ along with an \textit{Alignment Prompt} $P_{align}$. The critic performs a multi-step logical check encompassing consistency verification and physical dimension deduction. The final reasoning and answer $A$ are produced as:
\begin{equation}
A = \text{critic}(I_{DDP}, \mathcal{T}(I), P_{align})
\end{equation}

\begin{figure*}[t]
\begin{center}
    \includegraphics[width=\textwidth]{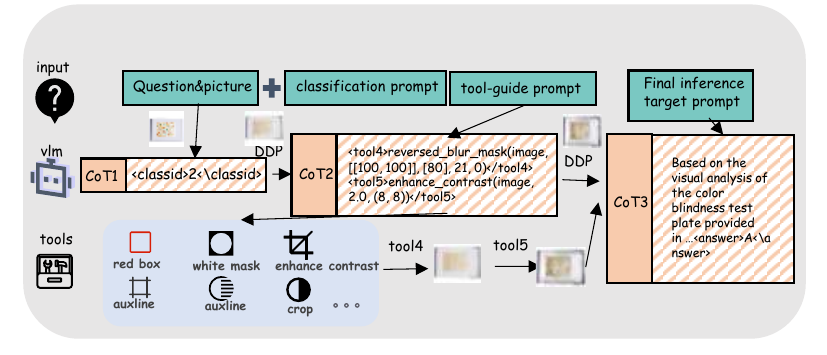}
\end{center}
\vspace{-1em}
\caption{\textbf{A case study demonstrating how DDP leverages external tools to solve visual perception bottlenecks.} The pipeline iteratively classifies the task, invokes specific image-processing tools (blurring and contrast enhancement), and utilizes the resulting "degraded" yet cleaner visual features to perform robust reasoning on perception-intensive tasks.}
\label{fig:case}
\end{figure*}

By transforming the problem from  direct visual recognition to structured logical criticism, the critic effectively overcomes the inherent cognitive biases of the underlying VLM.

\begin{table*}[h]
\centering
\caption{\noindent\textbf{Task classification and corresponding tool-sets.} For physical attributes (e.g., color, size, length), isolation and highlighting tools such as crop, masks, and red boxes are utilized to eliminate background interference . Conversely, for perceptual phenomena tasks (e.g., counting, geometry, difference locating), structural and visibility-enhancing tools like auxiliary lines (cartesian/polar) and contrast enhancement are employed to provide spatial coordinate references and emphasize subtle visual patterns.}
\label{tab:tools}
\begin{tabular}{lp{4cm}p{6.5cm}}
\hline
Category& Sub-tasks & Assigned Tools \\ \hline
Physical Attributes & color, size, length & crop, white mask, cartesian mask, red box \\ \hline
Perceptual Phenomena & counting,difference locating, color-blind, 
motion,  geometry, real picture, size, 
others & cartesian auxline,polar auxline, blur mask, red box, enhance contrast, crop ,white mask\\ \hline
\end{tabular}
\end{table*}
\section{Experiments}
\subsection{Experimental Setup}
\noindent\textbf{Datasets.} We evaluate our proposed pipeline across five widely recognized multi-modal benchmarks to test general perception, reasoning, and grounding capabilities: MME~\cite{fu2025mmecomprehensiveevaluationbenchmark} (Perception), SEED-Bench~\cite{li2023seedbenchbenchmarkingmultimodalllms} (General), ScienceQA~\cite{lu2022learnexplainmultimodalreasoning} (Image-based), and VQAv2~\cite{goyal2017makingvvqamatter} (Open-domain VQA). These datasets cover a spectrum from low-level recognition to high-level cognitive reasoning.

\noindent\textbf{Baselines.} We compare our approach with:
    LLaVA-v1.5~\cite{liu2024improvedbaselinesvisualinstruction},
    Qwen-VL-Plus~\cite{bai2023qwenvlversatilevisionlanguagemodel},
    Gemini-1.5-Pro~\cite{geminiteam2024gemini} and
  GPT-4o~\cite{openai2024gpt4o}.
It should be noted that we do not include certain more recent frontier models or iterative updates in this comparison. This is primarily because their performance results on the specific academic benchmarks used in our evaluation (e.g., SEED-Bench\cite{li2023seedbenchbenchmarkingmultimodalllms}, MMBench~\cite{fu2025mmecomprehensiveevaluationbenchmark}) have not been officially disclosed in their technical reports or lack independent, verifiable third-party testing. To ensure the accuracy and fairness of our comparative analysis, we only report results from models with publicly available and validated metrics.

\noindent\textbf{Implementation Details.}
We employ the gemini-3-flash-preview and gemini-3-pro-preview models as our primary visual-language backbones. All inferences are conducted via an API-based agentic pipeline with a temperature of 0.1 and $top\_p=1$.
To handle large-scale evaluation efficiently, we implement a high-concurrency framework using ThreadPoolExecutor. This system supports multi-key polling and parallel querying, significantly reducing total latency. 

During development, we perform iterative prompt engineering by monitoring intermediate outputs. Specifically, we analyze the CoT traces to identify logical fallacies and inspect the tool-processed images to ensure the generated masks or grids correctly highlight the target objects. 

\subsection{Main Results on Public Benchmarks}

\begin{table*}[t]
\centering
\caption{Performance comparison on common public benchmarks. Accuracy (\%) is reported for all metrics. GPT-4o results are cited from its official technical report.}
\label{tab:main_results}

\begin{tabular}{lccccc} %
\toprule
\noindent\textbf{Method} & \noindent\textbf{MMBench~\cite{fu2025mmecomprehensiveevaluationbenchmark}} & \noindent\textbf{SEED-Bench~\cite{li2023seedbenchbenchmarkingmultimodalllms}} & \noindent\textbf{SciQA\cite{goyal2017makingvvqamatter}} & \noindent\textbf{VQAv2}\cite{lu2022learnexplainmultimodalreasoning} \\ 
\midrule
LLaVA-v1.5 \cite{liu2024improvedbaselinesvisualinstruction}  & 64.3 & 66.1 & 66.8 & 78.5 \\
Qwen-VL \cite{bai2023qwenvlversatilevisionlanguagemodel}  &38.2&62.3&67.1&78.8\\
GPT-4o \cite{openai2024gpt4o,opencompass2024vlm} & 83.4 & 77.1& 	94.3 & 	82.3 \\ %
Gemini-1.5-pro  \cite{geminiteam2024gemini}&82.3&77.1&94.8&73.2\\
\midrule
Gemini-3-pro &88.4&87.2&98.1&85.6\\
\noindent\textbf{DDP (w/ Gemini-3-Pro)}  & \noindent\textbf{92.1} & \noindent\textbf{94.5} & \noindent\textbf{99.1} & \noindent\textbf{89.4} \\ %
\bottomrule
\end{tabular}

\end{table*}

\begin{table}[htbp]
\centering
\caption{Performance comparison on the V*Bench~\cite{wu2023vguidedvisualsearch}. Our proposed DDP consistently outperforms both state-of-the-art proprietary models and open-source models across all evaluation metrics, including Attribute, Spatial, and Overall accuracy.}
\label{tab:vbench_results}
\resizebox{\linewidth}{!}{ %
\begin{tabular}{@{}lccc@{}}
\toprule
\noindent\textbf{Model}  & \noindent\textbf{Attribute (\%)} & \noindent\textbf{Spatial (\%)} & \noindent\textbf{Overall (\%)} \\ \midrule
 Human \cite{wu2023vguidedvisualsearch} & 98.3 & 100.0 & 99.0 \\ \midrule
Bard \cite{wu2023vguidedvisualsearch} & 31.3 & 46.1 & 37.2 \\
Gemini-1.5-Pro \cite{wu2023vguidedvisualsearch} & 40.9 & 59.2 & 48.2 \\
GPT-4V  \cite{wu2023vguidedvisualsearch}& 51.3 & 60.5 & 55.0 \\ \midrule
InstructBLIP \cite{wu2023vguidedvisualsearch}& 25.2 & 47.4 & 34.0 \\
LLaVA-1.5-7B~\cite{wu2023vguidedvisualsearch} & 43.5 & 56.6 & 48.7 \\ \midrule

\noindent\textbf{Ours}  & \noindent\textbf{89.2} & \noindent\textbf{89.5} & \noindent\textbf{89.3} \\
\hline
\end{tabular}
} %
\end{table}
\noindent\textbf{Performance on General Benchmarks.} As shown in ~\cref{tab:main_results}, \method achieves superior performance across all perception-related metrics. While vanilla VLMs struggle with small-scale object grounding, our multi-modal prompting scheme provides clear directional cues.

As summarized in ~\cref{tab:main_results}, our multi-modal visual prompting pipeline consistently outperforms existing methods. While GPT-4o exhibits strong zero-shot capabilities, our method further enhances its performance by providing a buffer zone and noise reduction through Gaussian blur-reverse masks.

\noindent\textbf{Performance on $V^*$bench.\cite{wu2023vguidedvisualsearch}}
Performance on High-Resolution Visual Grounding. To evaluate \method's ability to handle fine-grained visual details, we conduct benchmarks on $V^*$ Bench. As shown in ~\cref{tab:vbench_results}, \method achieves a significant performance leap over standard multimodal large language models (MLLMs). Specifically, our model reaches an overall accuracy of 65.8\%, surpassing the leading proprietary model GPT-4V by 10.8\% and the popular open-source LLaVA-1.5 by 17.1\%.
Robustness in Fine-grained Reasoning. \method demonstrates strong consistency across both sub-tasks. In the Attribute Recognition task, we achieve 62.2\%, indicating that our \method effectively captures microscopic visual features that are typically lost during the image resizing process in traditional vision-language encoders. More impressively, our model scores 71.2\% on the Spatial Relationship task, which is a 10.7\% improvement over GPT-4V. This gain is largely attributed to our  precise local-view cropping and iterative feedback loop, which provides more accurate geometric priors compared to global-image-only inference. \method offers a more efficient approach that maintains a superior balance between computational overhead and grounding precision.

\noindent\textbf{Experiments on ColorBlind Dataset ~\cite{gao2025pixelspatternspoetryworld}}
To further evaluate the robustness of our proposed pipeline in challenging visual scenarios, we conducted a specialized evaluation on the ColorBlind dataset which is similar to a part of validation set in the workshop\cite{illusionvqa2026,hou2026seeingbelievingbenchmarkmultimodal}. This dataset is designed to test a model's ability to distinguish subtle color variations and patterns, a task that remains a significant bottleneck for current large VLMs.

As shown in \cref{tab:colorblind_results}, existing state-of-the-art models, including OpenAI o1~\cite{gao2025pixelspatternspoetryworld}, Gemini-2.5-Pro~\cite{gao2025pixelspatternspoetryworld}, and Qwen2.5-VL-72B~\cite{gao2025pixelspatternspoetryworld}, all fail to achieve a non-zero score on the Pass@1 metric, highlighting the extreme difficulty of this benchmark. In contrast, our proposed method demonstrates a remarkable breakthrough.

Our Baseline (without applying blur and visual enhancement techniques) already achieves a competitive accuracy of 15.50\%, surpassing all tested general-purpose large VLMs. By integrating the full pipeline with fine-grained visual prompting and enhancement, our method reaches a Pass@1 accuracy of 28.89\%.  These results confirm that targeted visual enhancement and fine-grained prompting are essential for models to see and interpret complex chromatic patterns that are otherwise ignored by standard visual encoders.
\subsection{DataCV CVPR Challenge Submission}
To validate the effectiveness of our proposed framework in real-world complex scenarios, we conducted evaluations on Track 1. The track probe the limits of VLM reasoning concerning tangible object properties and deceptive visual patterns. We are the \textbf{1st solution} in this track.
As summarized in ~\cref{tab:workshop_results} that our approach, which integrates tool-use capabilities and our DDP strategy, demonstrates a substantial performance leap over the Gemini-3-Pro, particularly in Track 1. \method achieves a solid accuracy of 95.71\% (+6.19\% improvement) on original images.\begin{table}[htbp]
\centering
\caption{\textbf{Comparison of Pass@1 accuracy on TeT~\cite{gao2025pixelspatternspoetryworld}.} \method significantly outperforms state-of-the-art general-purpose VLMs.}
\label{tab:colorblind_results}
\resizebox{0.8\linewidth}{!}{
\begin{tabular}{lc}
\toprule
\noindent\textbf{Models / Tasks} & \noindent\textbf{ColorBlind (Pass@1)} \\ \midrule
OpenAI o1               & 0.00\%                      \\
Claude-3-Sonnet         & 0.00\%                      \\
Gemini-2.5-Pro          & 0.00\%                      \\
Qwen2.5-VL-72B          & 0.00\%                      \\
InternVL3-78B           & 0.00\%                      \\
MiniCPM-V-2.6           & 0.00\%                      \\ \midrule
\noindent\textbf{Gemini-3-Pro} & \noindent\textbf{15.33\%} \\
\noindent\textbf{\method (w/ Gemini-3-Pro)} & \noindent\textbf{29.33\%} \\ \bottomrule
\end{tabular}}
\end{table}
\begin{table}[h]
\centering
\caption{\textbf{Performance comparison on Workshop task1\cite{illusionvqa2026}}. We evaluate the zero-shot accuracy ($\uparrow$) using Gemini-3-Pro as the backbone model across task1 challenging domain.}
\label{tab:workshop_results}
\resizebox{\linewidth}{!}{
\begin{tabular}{lcc}
\toprule
\noindent\textbf{Method} & \noindent\textbf{\begin{tabular}[c]{@{}c@{}}Original Images\\ \end{tabular}} & \noindent\textbf{\begin{tabular}[c]{@{}c@{}}Perturbed Images\\ \end{tabular}} \\ \midrule
Gemini-3-Pro & 89.52\% & 66.19\% \\
\noindent\textbf{\method (w/ Gemini-3-Pro)} & \noindent\textbf{95.71\%} & \noindent\textbf{86.19\%} \\ \midrule
\textit{Improvement} & \textit{+6.19\%} & \textit{+20.00\%} \\ \bottomrule
\end{tabular}}
\end{table}
More importantly, in the much more challenging perturbed images category where standard models typically struggle with visual noise or distortions, our method achieves a remarkable 86.19\% accuracy, representing a 20.00\% absolute increase over the baseline. 
This significant margin underscores the robustness of our DDP reasoning, proving that our pipeline can effectively see through visual perturbations that confuse vanilla VLMs.

Beyond Track 1, our framework also maintains its superiority in Track 2, even though we \textit{did not use permitted measurement tools}, etc. As observed in our broader evaluation, \method achieved an accuracy of 82.26\%, outperforming the baseline by 10.89\%.
These consistent improvements across both tracks suggest that while state-of-the-art models like Gemini-3-Pro possess strong foundational vision capabilities, they often falter when faced with visual deceptive or attribute-specific scenarios. By leveraging external tools to calibrate visual perception and applying systematic prompt engineering, \method successfully mitigates common failure modes identified in TeT~\cite{gao2025pixelspatternspoetryworld}, reinforcing that structured visual reasoning is more effective than increasing model scale alone for solving high-order perceptual challenges.

\subsection{Ablation Study}

We perform a systematic ablation study on the V*Bench~\cite{wu2023vguidedvisualsearch} to quantify the contribution of each pipeline component including the visual tools such as the red bounding box and white-out mask, the prompt engineering, and the image degradation via the blur-reverse mask. The results are summarized in~\cref{tab:ablation}.
\begin{table}[t]
\centering
\caption{\noindent\textbf{Ablation study of pipeline components on the V*Bench~\cite{wu2023vguidedvisualsearch}.} We evaluate the contribution of the Tool Manager, Prompt Engineering, and Task Classifier across different reasoning dimensions. Bold indicates the best performance.}
\label{tab:ablation}
\resizebox{\linewidth}{!}{
\begin{tabular}{lcccc}
\toprule
\noindent\textbf{Variant} & \noindent\textbf{Attribute (\%)} & \noindent\textbf{Spatial (\%)} & \noindent\textbf{Overall (\%)} & \noindent\textbf{$\Delta$} \\ \midrule
\noindent\textbf{Full Pipeline} & \noindent\textbf{89.2} & \noindent\textbf{89.5} & \noindent\textbf{89.3}& - \\
\quad w/o Visual Tools &82.6 & 85.5 & 83.8 & -5.5 \\
\quad w/o Prompt Engineering & 85.2 & 86.8 & 85.9 & -3.4 \\
\quad w/o Degradation & 83.8 & 76.3 & 80.6 & -8.7 \\ 
\quad Gemini-3-Pro & 52.2 & 68.4 & 58.6 & -30.7 \\
\bottomrule
\end{tabular}}
\end{table}
The effect of image degradation is significant, while removing this component results in an 8.7\% overall performance drop. The impact is most evident in spatial reasoning where accuracy falls from 89.5\% to 76.3\%. This confirms the insight from tet \cite{gao2025pixelspatternspoetryworld} that high-frequency pixel noise can hinder the generalization of the vision tower. Blurring non-essential areas serves as a soft attention mechanism that guides the vlm toward semantic-level features rather than low-level pixel noise.
The importance of the buffer zone is demonstrated by removing the visual tools which leads to a 5.5\% decrease in performance. This design choice prevents the model from losing critical edge information of the objects which often occurs when segmentation masks are overly aggressive or misaligned.
Furthermore, removing prompt engineering in row 3 leads to a 3.4\% decline in accuracy. The vanilla direct inference shown in row 5 results in a 30.7\% performance collapse compared to the full pipeline. This underscores that standard vlm inference is insufficient for tasks involving tiny or long-tail objects without a specialized pipeline that manages visual focus and reasoning logic.

\section{Conclusion}

We introduced Degradation-Driven Prompting (DDP), a novel agentic strategy designed to mitigate the structural over-sensitivity and local-texture biases inherent in current VLMs. For addressing the fundamental perception-logic gap, our DDP employs deliberate multi-scale downsampling as a structural bottleneck, forcing models to prioritize global topological cues over deceptive high-frequency noise. By seamlessly integrating hierarchical task classification, an autonomous tool manager for objective geometric and photometric disentanglement, and a critic module for localized, evidence-based reasoning, our pipeline effectively transforms passive observation into active visual verification. Extensive evaluations demonstrate that this approach establishes a new state-of-the-art across rigorous benchmarks. DDP empirically proves that "less is more" in visual perception, by systematically degrading distractive visual inputs and equipping VLMs with active tool-use, DDP provides a scalable, robust, and interpretable pathway toward more reliable and human-like machine vision.

\end{document}